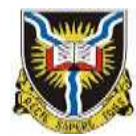

# A Predicting Phishing Websites Using Support Vector Machine and Multi-Class Classification Based on Association Rule Techniques

[1]Woods, Nancy C., [2]Agada, Virtue Ene and [3]Ojo, Adebola K.

[1]chyn.woods@gmail.com, [2]agadavirtue@gmail.com [3]adebola_ojo@yahoo.co.uk
[1,2,3]Department of Computer Science, University of Ibadan, Ibadan, Nigeria

**Abstract**

Phishing is a semantic attack which targets the user rather than the computer. It is a new Internet crime in comparison with other forms such as virus and hacking. Considering the damage phishing websites has caused to various economies by collapsing organizations, stealing information and financial diversion, various researchers have embarked on different ways of detecting phishing websites but there has been no agreement about the best algorithm to be used for prediction. This study is interested in integrating the strengths of two algorithms, Support Vector Machines (SVM) and Multi-Class Classification Rules based on Association Rules (MCAR) to establish a strong and better means of predicting phishing websites. A total of 11,056 websites were used from both PhishTank and yahoo directory to verify the effectiveness of this approach. Feature extraction and rules generation were done by the MCAR technique; classification and prediction were done by SVM technique. The result showed that the technique achieved 98.30% classification accuracy with a computation time of 2205.33s with minimum error rate. It showed a total of 98% Area under the Curve (AUC) which showed the proportion of accuracy in classifying phishing websites. The model showed 82.84% variance in the prediction of phishing websites based on the coefficient of determination. The use of two techniques together in detecting phishing websites produced a more accurate result as it combined the strength of both techniques respectively. This research work centralized on this advantage by building a hybrid of two techniques to help produce a more accurate result.

## I. INTRODUCTION

The word phishing from the phrase "website phishing" is a variation on the word "fishing". The idea is that bait is thrown out with the hopes that a user will grab it and bite into it just like the fish. In most cases, bait is either an e-mail or an instant messaging site, which will take the user to hostile phishing websites. [1]

Phishing is a new identity theft crime. The media report stories almost on a daily basis about an organization that has customers targeted by a phishing attack. While financial organizations try always to improve their security techniques in order to protect their customers, phishers develop even more sophisticated attacking techniques. Phishing websites are fake web pages that are created by malicious people to imitate web pages of real websites [2].

Phisher typically creates web pages that are visually very similar to the real web pages in order to scam their victims. An unaware client might be easily deceived by this kind of scam. The victims of a phishing web page may expose their bank account, password, credit card number, or other important information to the phishing web page owners. While phishing is a relatively new internet crime when compared to other forms (for example, viruses and hacking), a recognizable increase in the number and severity of phishing attacks is reported [3]. Considering the damage phishing websites have caused to various economy by collapsing organizations, stealing information, financial diversion, as well as making organizations, businesses and individuals to run bankrupt, various researchers have embarked on different ways of detecting these phishing websites but there is no agreement among researchers about the best algorithm to be used for prediction. This is born out of the variations of strengths of various algorithms for prediction, because some algorithms are good at classification while some are good at prediction. On this premise, this study is interested in collapsing the strengths of two algorithms to establish a stronger and better means of predicting phishing websites.

## II. LITERATURE REVIEW

### A. A Conceptual Review of Phishing

The United States Computer Emergency Readiness Team (US-CERT) defines phishing as a form of social engineering that uses email or malicious websites (among other channels) to solicit personal information

Woods, N. C., Agada, V. E. and Ojo, A. K. (2018). A Predicting Phishing Websites Using Support Vector Machine and Multi-Class Classification Based on Association Rule Techniques, *University of Ibadan Journal of Science and Logics in ICT Research (UIJSLICTR)*, Vol. 2, No. 1, pp. 28 - 39
©U IJSLICTR    Vol. 2, No. 1 June   2018

from an individual or company by posing as a trustworthy organization or entity. Phishing attacks often use email as a vehicle, sending email messages to users that appear to be from an institution or company that the individual conducts business with, such as a banking or financial institution, or a web service through which the individual has an account. [4]

The goal of a phishing attempt is to trick the recipient into taking the attacker's desired action, such as providing login credentials or other sensitive information. For instance, a phishing email appearing to come from a bank may warn the recipient that their account information has been compromised, directing the individual to a website where their username and/or password can be reset. This website is also fraudulent, designed to look legitimate, but exists solely to collect login information from phishing victims. [4]

### B. Phishing Vectors

According to [5], Phishing vectors are the routes that malicious attacks may take to get past your defenses and infect your network. He spoke on only six Phishing vectors in particular namely:
- Network – The perimeter of your network, usually protected by something like a firewall.
- User – Attackers often use social engineering and social networking to gather information and trick users into opening a pathway for an attack into a network.
- Email – Phishing attacks and malicious attachments target the email threat vector.
- Web Application – SQL Injection and Cross-Site Scripting are just two of the many attacks that take advantage of an inadequately protected Web Application threat vector.
- Remote Access – A corporate device using an unsecured wireless hotspot can be compromised and passed on to the corporate network.
- Mobile – Smart phones, tablets, and other mobile devices can be used as devices to pass malware and other attacks on to the corporate network. Additionally, mobile malware may be used to steal useful data from the mobile device.

### C. REVIEW OF RELATED WORKS

Ajlouni, *et. al.,* [6] carried out a study titled Detecting Phishing Websites Using Associative Classification (MCAR, CBA). This paper's main goal was to investigate the potential use of automated data mining techniques in detecting the complex problem of phishing Websites in order to help all users from being deceived or hacked by stealing their personal information and passwords leading to catastrophic consequences. Experimentations against phishing data sets and using different common associative classification algorithms (MCAR and CBA) and traditional learning approaches was conducted with reference to classification accuracy. However, CBA and MCAR are both associative classifiers. Hence it generates a very large number of association classification rules.

Suganya [7] did a review on Phishing Attacks and various Anti Phishing techniques. However, He only discussed the various types of phishing techniques and phishing attacks but did not implement the techniques to solve the phishing attacks.

Leena and Er. Amrit [8] did a research on Detecting of Phishing Websites using SVM Technique. However, they did not explore the topic on a collected data set, neither did they propose an algorithm to classify the websites as legitimate or non-legitimate.

Kadam, and Pawar [9] did a study on Comparison of Association Rule with Pruning and Adaptive technique for classification of phishing dataset. However, there is need to modify the Adaboost to improve accuracy and speed.

Thabtah, *et. al.,* [10] conducted a research on MCAR: Multi-class Classification based on Association Rule. In this work, a new classification method called multi-class classification based on association rules (MCAR) was presented. However, there was no extraction of multiple class labels using association rule discovery.

### III. METHODOLOGY

#### A. Implementing Support Vector Machines (SVM) Algorithm

- Equation of n-dimensional hyperplane can be written as $W^T X = C$, where $W= [w_1, w_2, \ldots \ldots w_n]$ and $X=[X_1, X_2, \ldots \ldots X_n]$
- Hyperplane separates the space into two half spaces (positive half space and negative half space).
- A hyperplane is also known as linear discriminant as it linearly divides the space in two halves.
- Support vector machine is a linear discriminant.

Therefore, the equation of a hyperplane can be written as $W^T X = C$, some properties of W and C are:

i. C determines the position of hyperplane and W determines the orientation (angle with axis) of a hyperplane. How? let's analyze it by taking a 2-dimensional surface, for a two dimensional surface hyperplane will be a line and the equation will be $w_1 x_1 + w_2 x_2 = C$, this equation can be written as $x_2 = -(w_1/w_2) * x_1 + C$, now compare it with general line equation $y = mx + C$, as you can see m determines the angle with the axis $(-w_1/w_2)$ and C determines position in the X-Y Plane.

ii. Vector W is orthogonal to the hyperplane, the direction of W is in the direction of positive half. How? Let's take two points m and n in hyperplane, now these points are in hyperplane so they will

29

satisfy the equation $W^T m = C$ and $W^T n = C$. Subtracting these two equations we will get $W^T(m-n) = 0$, now direction of vector (m-n) will be in the direction of plane and as the dot product of W with (m-n) is zero, it means vector W is orthogonal (perpendicular in layman term) to hyperplane.

iii. Shortest distance between a point and the hyperplane;

To find the minimum distance between Point X and decision boundary, we need to find point Xp in decision boundary such that the vector (Xp-X) is orthogonal to the boundary. This becomes an optimization problem with an objective function:

Find Xp Such that ||Xp-X|| is minimum and $W^T Xp = C$ (as Xp is on decision boundary).

Solving above optimization problem requires Formulation of Lagrangian and applying Karush-Kuhn-Tucker (KKT) conditions. For the sake of simplicity, optimization result is produced.

$Xp = X - ((W^T X - C)W/||W||^2)$ and the distance D between Xp and X

$D = (W^T X - C)/||W||$ .... (*Equation 1*)

The above Distance equation is very important as it forms the basis of Support Vector Machines (SVM).

### B. Multi-Class Classification Based on Association Rule (MCAR) Algorithm

This is a new classification method. It is a recently developed associative classification algorithm.

MCAR uses an efficient technique for discovering frequent items and employs a rule ranking method which ensures detailed rules with high confidence are part of the classifier.

The algorithm proposed consists of two phases:

i. **Rules Generation**: MCAR scans the training data set to discover frequent single items, and then recursively combines the items generated to produce items involving more attributes. In the first phase, MCAR scans the training data set to discover frequent single items, and then recursively combines the items generated to produce items involving more attributes. MCAR then generates ranks and stores the rules.

ii. **Classifier Builder:** In the second phase, the rules are used to generate a classifier by considering their effectiveness on the training data set.

**The main contributions of the MCAR approach:**
i. MCAR discovers and generates frequent items and rules in one phase.
ii. MCAR introduces a rule ranking technique that minimizes the use of randomization when a choose point must be made between two or more rules.

### C. MCAR-SVM ALGORITHM

Developing model of MCAR and SVM involves the merging of the strengths of the two algorithms which involves three specific stages such as rule building, pruning down of multiple rules and the rules generated will be used to find a hyperplane to identify the largest margin within the classes of result that is the best prediction. MCAR rule discovery method requires only one single data scan and then performs simple intersection between the TIDLists of ruleitems of size N-1 to generate candidate ruleitems of size N. Once all frequent ruleitems are discovered, MCAR algorithm generates the subset of those which hold larger confidence than the minimum confidence threshold as rules. When all rules are generated then the algorithm applies a ranking procedure to favour rules over each other. The basis of this rule favouring procedure is mainly the confidence value, and then support value and lastly the size of the rules (number of attributes values in the rule body). If two or more rules having similar confidence, support and rule size then the rank will be random.

Once all rules are sorted, then MCAR uses the database coverage pruning to remove redundant rules from taking any role in the prediction step. The outputs of the pruning are the subset of rules that are highly predictive and those represent the classifier. SVM is used to find the best hyperplane that is capable of creating the largest margin among the classes of response (legitimate and phishing websites). Once the classifier is produced, its predictive power is tested using cross validation or on test data set.

### THE MCAR-SVM MODEL STEPS

Step 1: The discovery of all frequent ruleitems.

Step 2: The production of all CARs that have confidences above the minconf threshold from frequent ruleitems extracted in Step 1.

Step 3: The selection of one subset of CARs to form the classifier from those generated at Step 2.

Step 4: Support Vector Machines come in to find a hyperplane that can separate two distinct sets of classes. This means that there is a weight vector *w* and a threshold *b'*, so that all positive training examples are on one side of the hyperplane, while negative training examples lie on the other side.

**D. OVERVIEW OF THE PROPOSED MODEL**

The main steps of the proposed model are as follows:
a) Website feature extraction
b) Using MCAR algorithm, in order to generate rules and rule pruning.
c) Prediction of result using SVM

**THE MODEL FOR EXTRACTION AND EVALUATION OF CHOSEN FEATURES**

The model used extracts features from the phishing page. Figure 1 shows the features and they are divided into four main categories depending on the impact factor in predicting phishing and then are allocated to one of the four categories. The proposed model uses output from higher level as the input to the next lower level, and then the MCAR mining technique is applied on the extracted features to generated rules. These rules are used for future classification and prediction of the websites.

**SYSTEM ARCHITECTURE**

The System is divided into 4 Modules as seen in Figure 2:

i. **Data Source/ Data Base Module**: This Module maintains data in the form of data sets. These datasets are normalized and filtered to get pure data without outliers. Feature contents of the data set are also extracted in this module. The data is also divided into training and testing data.
ii. **Association Rule Generation Module:** This module performs multi-class association rule mining and generates frequent item sets and generates association rules.
iii. **Classification Module:** This Module reads the rules produced in the classification module and classifies the data based on the rule as either legitimate or non-legitimate.
iv. **Performance Analysis Module**: This Module computes time complexity, space complexity, accuracy, error rate and precision on a number of classes and also evaluates the algorithm with another technique.

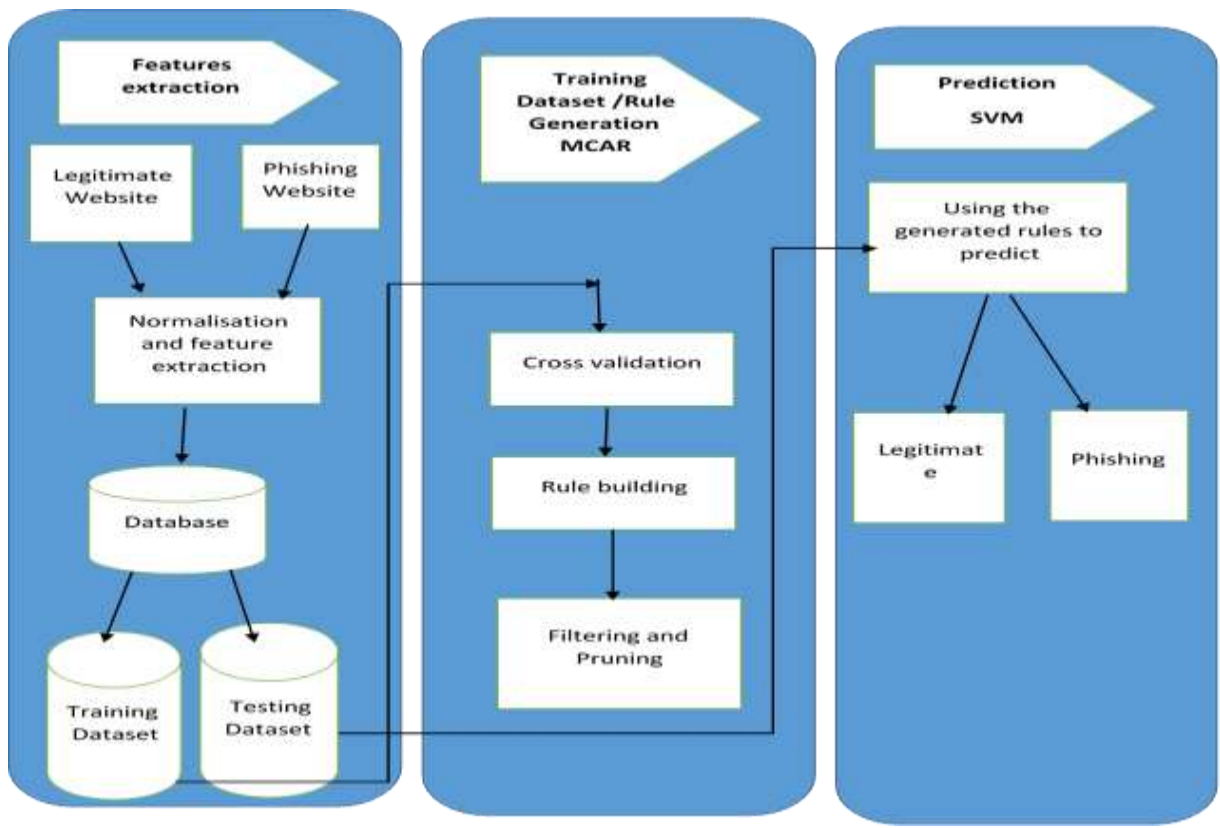

**Figure 1:** Proposed MCAR-SVM system for feature subset selection and classification Methodology

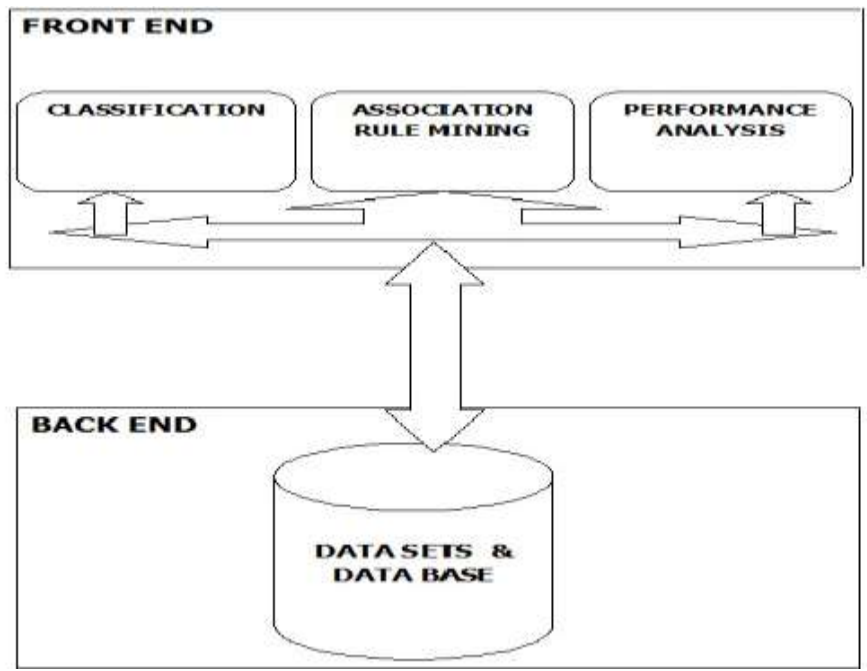

**Figure 2:** System Architecture

**D. FEATURE SELECTION**

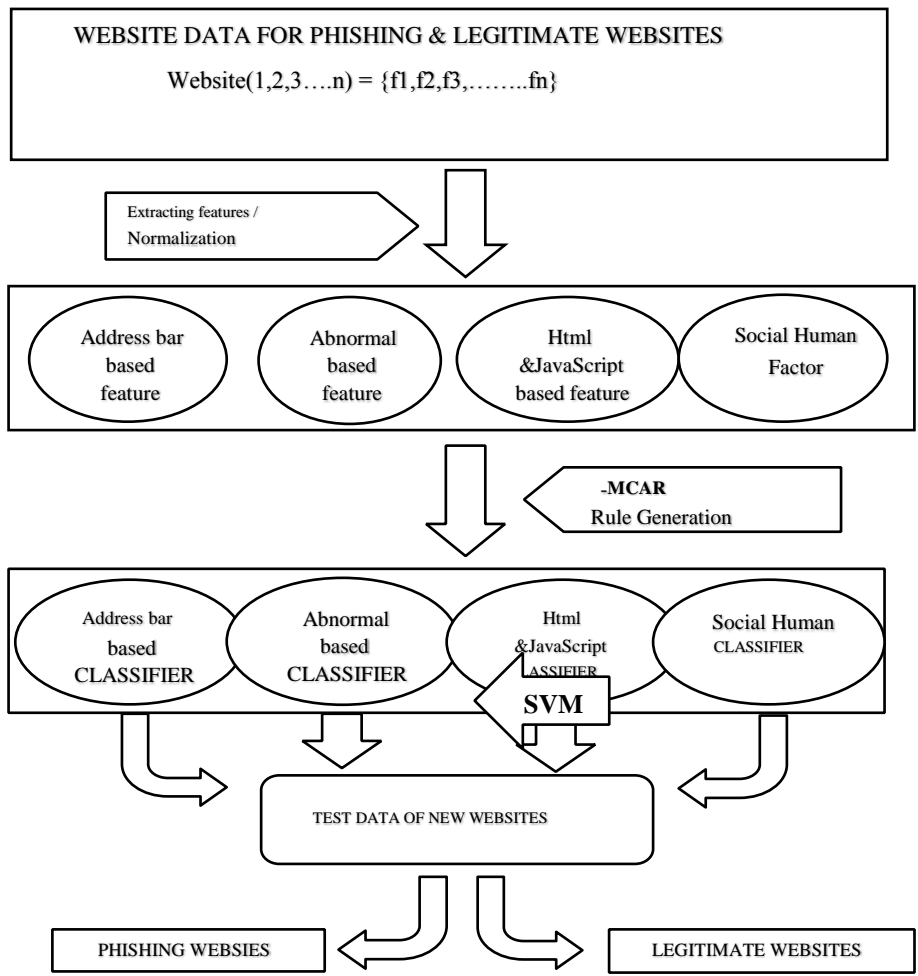

**Figure. 3** Feature extraction and phishing detection model

## IV. IMPLEMENTATION AND RESULTS

### A Datasets for Phishing websites

Figure 4 consists of the pre-processed phishing datasets that is been used for this study. It is saved in a .csv format.

**Figure 4:** Phishing Dataset

**Figure 5:** The start of website dataset loading by MCAR-SVM

**B. DESCRIPTION OF MCAR-SVM CLASSIFIER IMPLEMENTATION**

The MCAR-SVM system for website classification is executed by typing the file name "Training Data.csv" on the R console. The "Training Data.csv" initiates the execution of the MCAR-SVM by loading the pre-processed dataset. This is depicted in Figure 5.

Figure 6 reveals that the three-level process on the R console: the training, validating and testing processes of the integrated MCAR-SVM for classification of phishing website data set.

On the completion of the training process, the test_data and test_label were introduced to determine the effectiveness of this system. MCAR uses an efficient technique for discovering frequent items and employs a rule ranking method which ensures detailed rules with high confidence are part of the classifier. It scans the training data set to discover frequent single items, and then recursively combines the items generated to produce items involving more attributes. In the first phase, MCAR scans the training data set to discover frequent single items, and then recursively combines the items generated to produce items involving more attributes. MCAR then generates ranks and stores the rules. After pruning of rules, some features were also removed. Afterwards, the remaining viable rules after training were suggested to SVM in the prediction of Phishing websites.

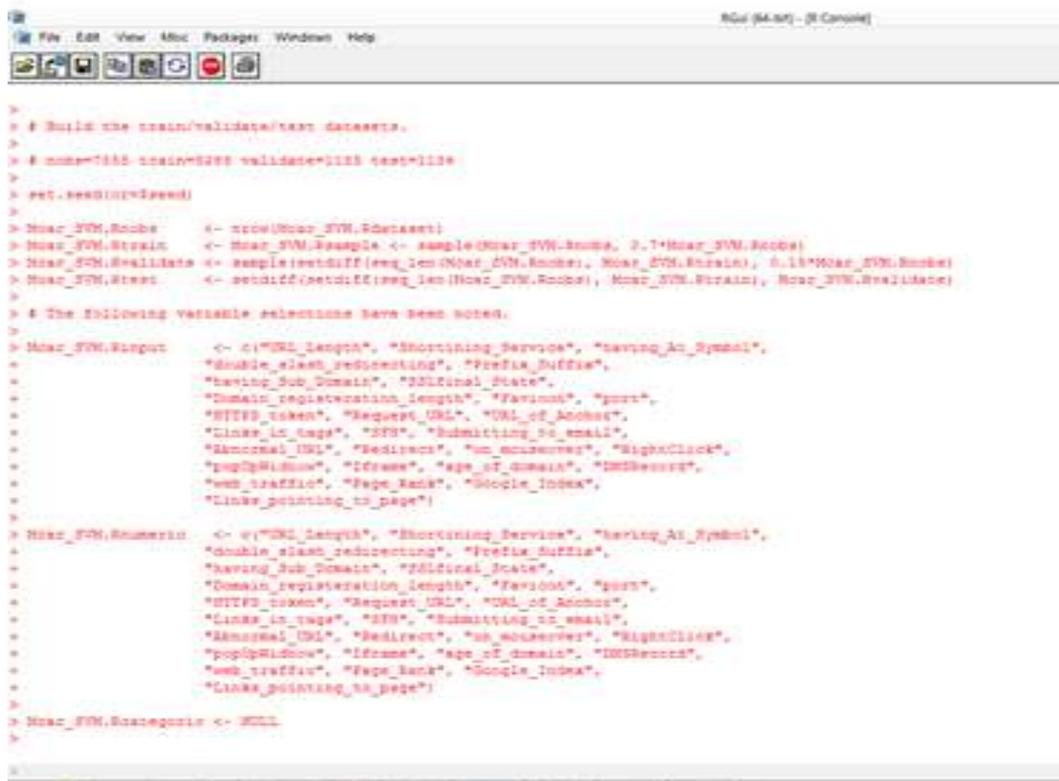

**Figure 6**: MCAR-SVM running to generate newer futures and rules

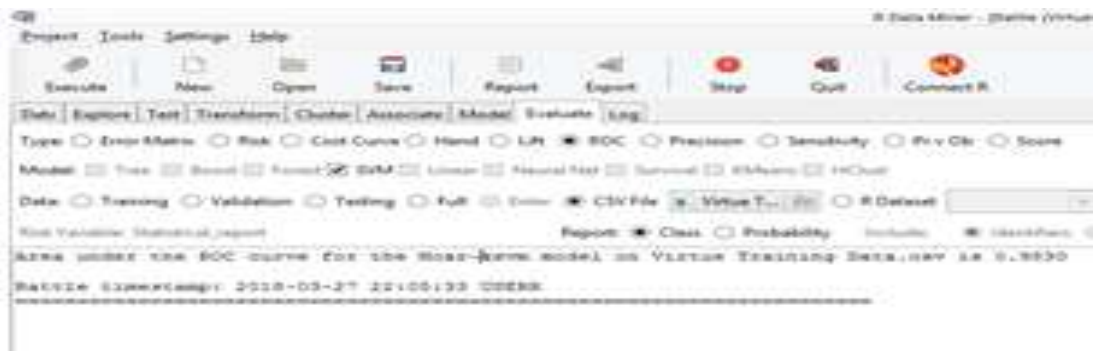

**Figure 7**: The result obtained using MCAR-SVM

The phishing website dataset used contained 11,056 websites, and 5005 were collected from Phishtank archive. The accuracy of the dataset classified using the combined MCAR-SVM is shown in Figure 7. The elapse time for the classification is also displayed. The MCAR-SVM accuracy is 98.30% while the elapse time is 2205.33 seconds.

The SVM algorithm was used on the website dataset classification by typing the file name "Virtue Phishing data.csv" to reload the dataset back to the workspace as shown in Figure 8.

The dataset are displayed in Figure 9 and afterwards trained and tested using SVM. After the data has been trained, the trained data were used on testing data to predict phishing from legitimate websites.

Figure 10 shows the computation time and the accuracy of the SVM classifier. 7,739 websites were used to train the SVM classifier and 3,317 websites were used to test it. The accuracy and the computation time obtained were 98.59% and 1515.02 seconds respectively for SVM.

**DESCRIPTION OF SVM CLASSIFIER IMPLEMENTATION**

**Figure 8:** The start of website dataset loading by SVM

**Figure 9**: The features of the dataset.

**Figure 10:** The result obtained using SVM.

35

**Table 1:** The summary of results obtained using SVM, Decision Tree and MCAR-SVM

| Algorithm | Classification Accuracy | Computation Time |
|---|---|---|
| SVM | 99.1% | 1515.02s |
| MCAR-SVM | 98.30% | 2205.33 s |
| Decision Tree | 90.85% | 1551.00s |

Table 1 shows the summary of results obtained using SVM, Decision Tree and MCAR-SVM. Considering the results obtained after the evaluation of SVM, Decision Tree and MCAR-SVM as shown in the Table 1, SVM has proven to yield higher classification accuracy of 99.1% within a lesser computational time of 1515.02s, followed by MCAR-SVM yields a classification accuracy of 98.30% within a computation time of 2205.33s, while Decision Tree yields a classification accuracy of 90.85% within a computational time of 1551.00s. This shows that on the R programming console the MCAR-SVM algorithm did not yield a better classification accuracy than SVM and Decision tree within reasonable computation time.

**Table 2:** The comparison of weighted average for different classifier SVM and MCAR-SVM.

| Algorithm | TP Rate | FP Rate | Precision | Recall | Error Rate |
|---|---|---|---|---|---|
| SVM | 0.530 | 0.033 | 0.788 | 0.880 | 0.05 |
| MCAR-SVM | 0.534 | 0.041 | 0.902 | 0.902 | 0.06 |
| Decision tree | 0.505 | 0.046 | 0.800 | 0.830 | 0.23 |

Table 2 showing the summary of comparison of weighted average for different classifier SVM and MCAR-SVM. It reveals that MCAR-SVM and the SVM algorithm displayed similar True positive rate although there are slight variances in their false positive rate and error rate, which is better than that of the decision tree. That is, MCAR-SVM correctly classified 53.4% of the websites as phishing, followed by SVM by 53%, while decision tree by 50.5%. On the other hand, SVM wrong classification of phishing websites was the least, by 3.3%, MCAR-SVM by 4.1% and Decision tree by 4.6%. By implication, SVM appeared superior in the classification of websites status (phishing or legitimate).

**IMPLICATION OF THE RESULTS OBTAINED**
SVM and Decision tree classifiers have been observed to consume a lot of computation resources and result in inaccurate classification in the face of a large website dataset using MATLAB commands. The result indicates that on the R programming console SVM appeared efficient than the proposed MCAR-SVM classifier a remarkable classification accuracy and computation time.

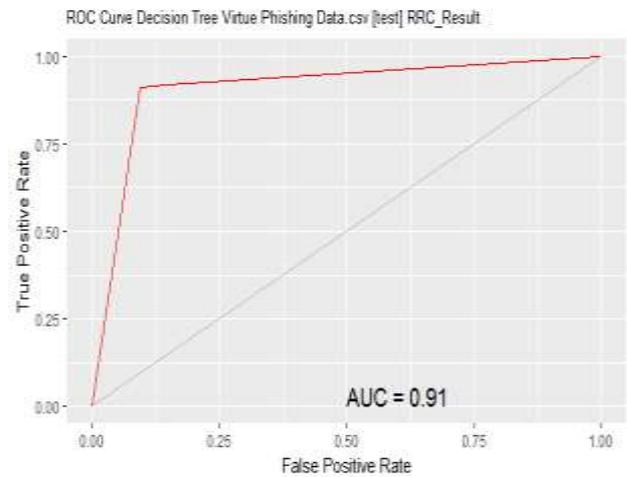

**Figure 11:** Receiver operation characteristics (ROC) showing 91% Area Under the Curve (AUC) proportion of accuracy in classification of phishing websites using decision tree.

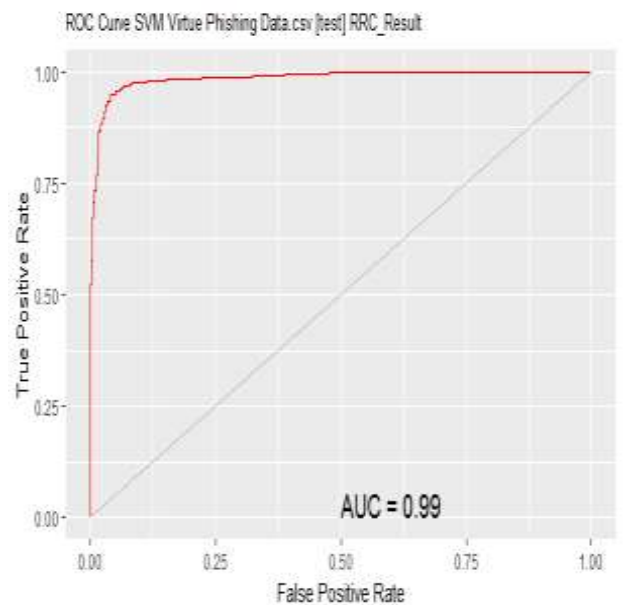

**Figure 12:** Receiver operation characteristics (ROC) showing 99% Area Under the Curve (AUC) proportion of accuracy in classification of phishing websites using SVM.

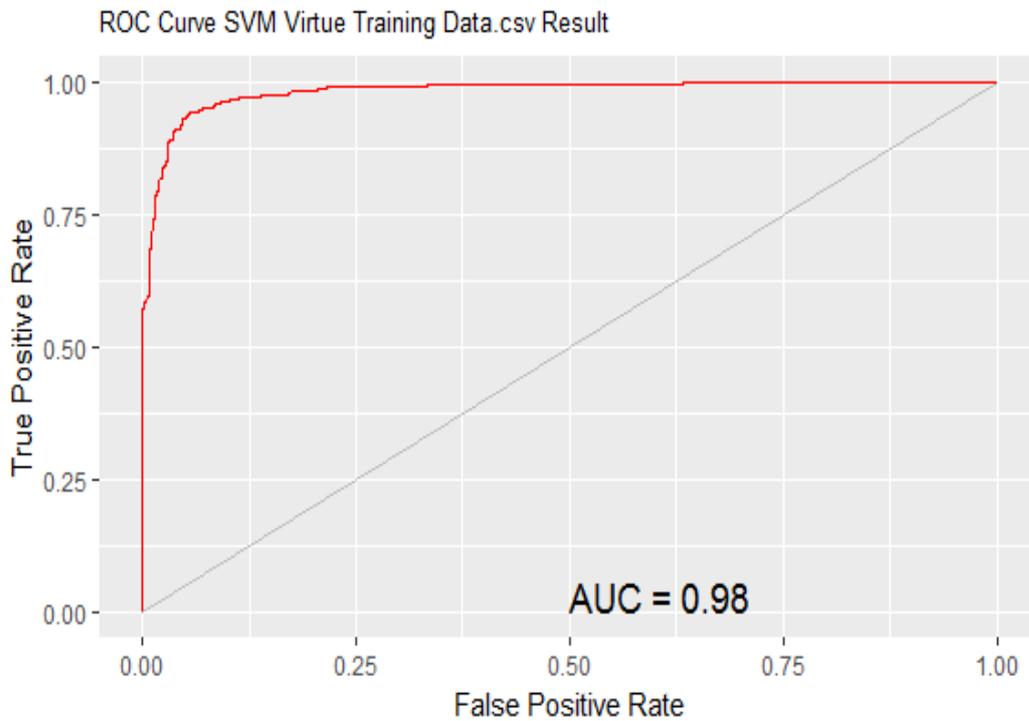

**Figure 13:** Receiver operation characteristics (ROC) showing 98% Area Under the Curve (AUC) proportion of accuracy in classification of phishing websites using MCAR-SVM.

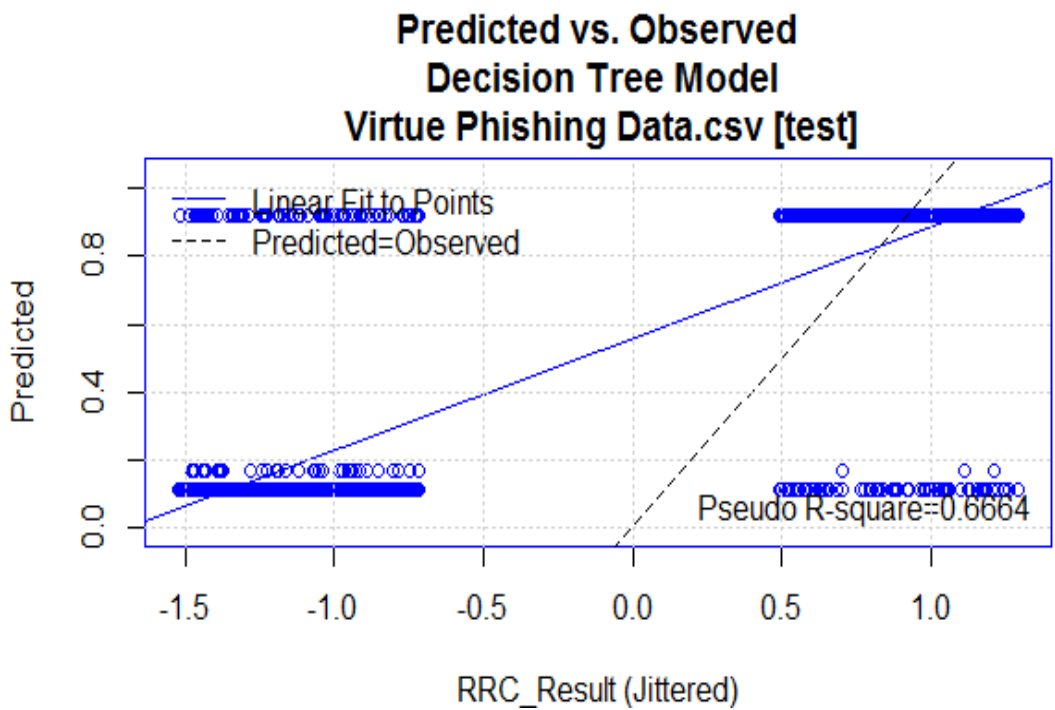

**Figure 14**: Decision tree model showing 66.64% (Pseudo R-square= 0.6664) variance in the prediction of phishing websites.

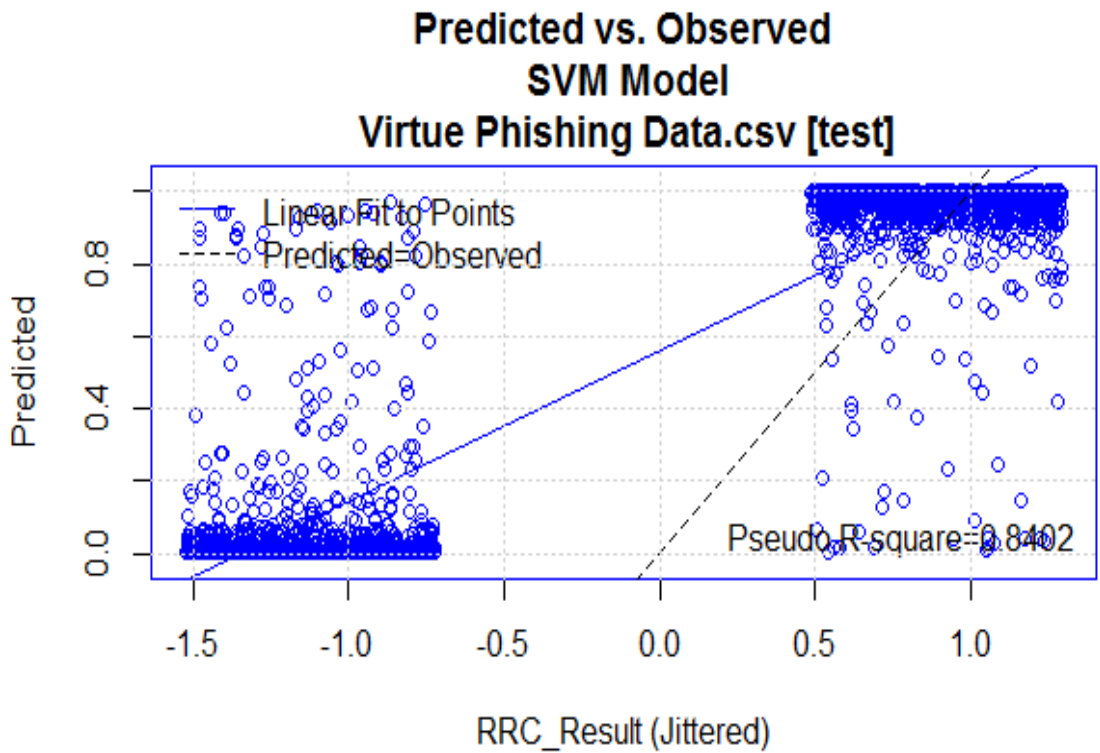

**Figure 15:** SVM model showing 84.02% (Pseudo R-square= 0.8402) variance in the prediction of phishing websites.

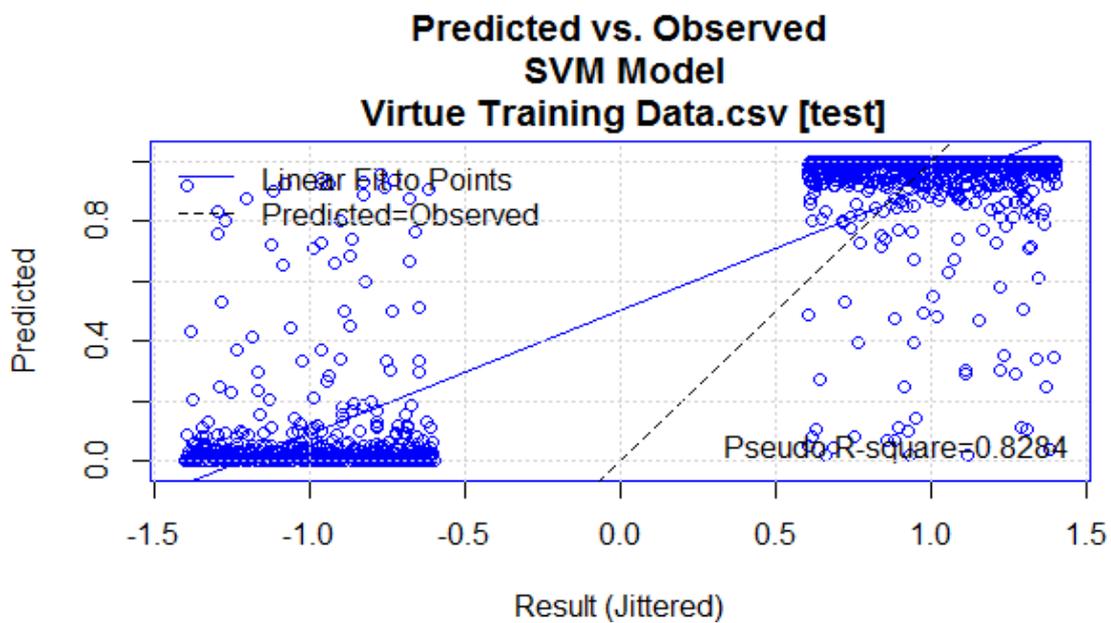

**Figure 16**: MCAR-SVM model showing 82.84% (Pseudo R-square= 0.8284) variance in the prediction of phishing websites.

## V. CONCLUSION

In this study, a heuristic-based phishing detection technique that employs multiple features of phishing sites was proposed. The method combines URL-based features used in previous studies with new features by analyzing phishing sites. Additionally, we generated an integration of two machine learning algorithms namely MCAR and SVM. It showed a high accuracy of 98.30% and a low error rate. However, SVM Technique achieved a higher accuracy in classification with 99.1% accuracy and a lesser computation time than the combined model (MCAR-SVM). The proposed technique can also provide security for personal information and reduce damage caused by phishing attacks because it can detect new and temporary phishing sites that evade existing phishing detection techniques.